\documentclass{article}

    \PassOptionsToPackage{numbers, compress}{natbib}

    \usepackage[preprint]{neurips_2020}

\usepackage[utf8]{inputenc} 
\usepackage[T1]{fontenc}    
\usepackage{hyperref}       
\usepackage{url}            
\usepackage{booktabs}       
\usepackage{amsfonts}       
\usepackage{nicefrac}       
\usepackage{microtype}      

\usepackage{graphicx}
\usepackage{subfigure}
\usepackage{amsmath}
\usepackage{bbm, dsfont}

\usepackage[algo2e]{algorithm2e}
\usepackage{algorithm}

\usepackage{xcolor}
\usepackage{xspace}

\newcommand{\mymath}[1]{\ensuremath{#1}\xspace}

\definecolor{myred}{rgb}{0.8,0,0}
\definecolor{mygreen}{rgb}{0,0.6,0}
\definecolor{myblue}{rgb}{0,0,0.7}

\newcommand{\reals}{\mymath{\mathbb R}}

\newcommand{\mcts}{{\sc mcts}\xspace}
\newcommand{\uvfas}{{\sc uvfa}s\xspace}
\newcommand{\uvfa}{{\sc uvfa}\xspace}
\newcommand{\ddpg}{{\sc ddpg}\xspace}
\newcommand{\mddpg}{{\sc m-ddpg}\xspace}
\newcommand{\her}{{\sc her}\xspace}

\newcommand{\curious}{{\sc curious}\xspace}
\newcommand{\anx}{AlphaNPI-X\xspace}
\newcommand{\anpi}{AlphaNPI\xspace}
\newcommand{\npi}{NPI\xspace}
\newcommand{\lstm}{{LSTM}\xspace}
\newcommand{\goalgen}{\Theta}  
\newcommand{\abm}{\Omega_{w}}  
\newcommand{\atomp}{\pi_{\theta}^{atomic}}  
\newcommand{\precond}{\phi}  
\newcommand{\postcond}{\psi}  

\title{Learning Compositional Neural Programs for Continuous Control}

\author{Thomas Pierrot\\
    InstaDeep\\
    \texttt{t.pierrot@instadeep.com}\\
    \And 
    Nicolas Perrin\\
    CNRS, Sorbonne Universit\'{e}\\
    \texttt{perrin@isir.upmc.fr}\\
    \And Feryal Behbahani\\
    DeepMind\\
    \texttt{feryal@google.com}\\
    \And 
    Alexandre Laterre\\
    InstaDeep\\
    \texttt{a.laterre@instadeep.com}\\
    \And 
    Olivier Sigaud\\
    Sorbonne Universit\'{e}\\
    \texttt{olivier.sigaud@upmc.fr}\\
    \And
    Karim Beguir\\
    InstaDeep\\
    \texttt{kb@instadeep.com}\\
    \And
    Nando de Freitas\\
    DeepMind\\
    \texttt{nandodefreitas@google.com}\\
}

\begin{document}

\maketitle

\begin{abstract}
We propose a novel solution to challenging sparse-reward, continuous control problems that require hierarchical planning at multiple levels of abstraction. Our solution, dubbed AlphaNPI-X, involves three separate stages of learning. First, we use off-policy reinforcement learning algorithms with experience replay to learn a set of atomic goal-conditioned policies, which can be easily repurposed for many tasks. Second, we learn self-models describing the effect of the atomic policies on the environment. Third, the self-models are harnessed to learn recursive compositional programs with multiple levels of abstraction. The key insight is that the self-models enable planning by imagination, obviating the need for interaction with the world when learning higher-level compositional programs. To accomplish the third stage of learning, we extend the AlphaNPI algorithm, which applies AlphaZero to learn recursive neural programmer-interpreters. We empirically show that AlphaNPI-X can effectively learn to tackle challenging sparse manipulation tasks, such as stacking multiple blocks, where powerful model-free baselines fail.
\end{abstract}

\section{Introduction}

Deep reinforcement learning (RL) has advanced many control domains, including dexterous object manipulation \citep{akkaya2019solving,andrychowicz2018learning,nagabandi2019deep}, agile locomotion \citep{tan2018sim} and navigation \citep{faust2018prm,held2018automatic}. 
Despite these successes, several key challenges remain. 
Stuart Russell phrases one of these challenges eloquently in \cite{russell2019human}: ``\emph{Intelligent behavior over long time scales requires the ability to plan and manage activity hierarchically, at multiple levels of abstraction}'' and ``\emph{the main missing piece of the puzzle is a method for constructing the hierarchy of abstract actions in the first place''}. If achieved, this capability would be ``\emph{The most important step needed to reach human-level AI}'' \cite{russell2019human}. This challenge is particularly daunting when temporally extended tasks are combined with sparse binary rewards. In this case the agent does not receive any feedback from the environment while having to decide on a complex course of action, and receives a non-zero reward only after having fully solved the task. 

Low sample efficiency is another challenge. In the absence of demonstrations, model-free RL agents require many interactions with the environment to converge to a satisfactory policy \citep{akkaya2019solving}. We argue in this paper that both challenges can be addressed by learning models and compositional neural programs.
In particular, planning with a learned internal model of the world reduces the amount of necessary interactions with the environment. By imagining likely future scenarios, an agent can avoid making mistakes in the real environment and instead find a sound plan before acting on the environment. 

Many real-world tasks are naturally decomposed into hierarchical structures. We hypothesize that learning a variety of skills which can be reused and composed to learn more complex skills is key to tackling long-horizon sparse reward tasks in a sample efficient manner. Such compositionality, formalised by hierarchical RL (HRL), enables agents to explore in a temporally correlated manner, improving sample efficiency by reusing previously trained lower level skills. Unfortunately, prior studies in HRL typically assume that the hierarchy is given, or learn very simple forms of hierarchy 
 in a model-free manner.

We propose a novel method, \emph{\anx}, to learn programmatic policies which can perform hierarchical planning at multiple levels of abstraction in sparse reward continuous control problems.
We first train low-level atomic policies that can be recomposed and re-purposed, represented by a \emph{single} goal-conditioned neural network. We leverage off-policy reinforcement learning with hindsight experience replay \citep{HER} to train these efficiently.
Next, we learn a transition model over the effects of these atomic policies, to imagine likely future scenarios, removing the need to interact with the real environment. 
Lastly, we learn recursive compositional programs, which combine low-level atomic policies at multiple levels of hierarchy, by planning over the learnt transition models, alleviating the need to interact with the environment.
This is made possible by extending the \anpi algorithm \cite{AlphaNPI} which applies AlphaZero-style planning \citep{silver2017mastering} in a recursive manner to learn recombinable libraries of symbolic programs. 

We show that our agent can learn to successfully combine skills hierarchically to solve challenging robotic manipulation tasks through look-ahead planning, even in the absence of any further interactions with the environment and where powerful model-free baselines struggle to get off the ground.\footnote{Videos of agent behaviour are available at: \url{https://sites.google.com/view/alphanpix}}

\section{Related Work}

Our work is motivated by the central concern of constructing a hierarchy of abstract actions to deal with multiple challenging continuous control problems with sparse rewards. In addition, we want these actions to be reused and recomposed across different levels of hierarchy through planning with a focus on sample efficiency. This brings several research areas together, namely multitask learning, hierarchical reinforcement learning (HRL) and model-based reinforcement learning (MBRL).

As we have shown, learning continuous control from sparse binary rewards is difficult because it requires the agent to find long sequences of continuous actions from very few information. Other works tackling similar  block stacking problems \cite{curiosity_driven_multi_criteria,li2019towards} have either used a very precisely tuned curriculum, auxiliary tasks or reward engineering to succeed. In contrast, we show that by relying on planning even in absence of interactions with the environment we can successfully learn from raw reward signals.

Multitask learning and the resulting transfer learning challenge have been extensively studied across a large variety of domains, ranging from supervised problems, to goal-reaching in robotics or multi-agent settings  \citep{taylor2009transfer,zhang2017survey}. 
A common requirement is to modify an agent's behaviour depending on the currently inferred task. 
Universal value functions \citep{uvfa} have been identified as a general factorized representation which is amenable to goal-conditioning agents and can be trained effectively \citep{barreto2019option}, when one has access to appropriately varied tasks to train from. Hindsight Experience Replay (\her) \citep{HER} helps improve the training efficiency by allowing past experience to be relabelled retroactively and has shown great promises in variety of domains. However, using a single task conditioning vector has its limitations when addressing long horizon problems. Other works explore different ways to break long-horizon tasks into sequences of sub-goals \citep{nair2019hierarchical, bharadhwaj2020dynamics}, or leverage \her and curiosity signals to tackle similar robotic manipulation tasks \citep{curiosity_driven_multi_criteria}. In our work, we leverage \her to train low-level goal-conditioned policies, but additionally learn to combine them sequentially using a program-guided meta-controller.

The problem of learning to sequence sub-behaviours together has been well-studied in the context of HRL \citep{sutton-options-original,option-critic} and has seen consistent progress throughout the years \citep{levy2017hierarchical,levy2018hierarchical,vezhnevets2017feudal,data-efficient-hrl}.
Learning both which sub-behaviours could be useful and how to combine them still remains a hard problem, which is often addressed by pre-learning sub-behaviours using a variety of signals (e.g. rewards \citep{barreto2019option}, demonstrations \citep{gupta2019relay}, state-space coverage \citep{eysenbach2018diversity,pong2019skew,lee2019efficient,islam2019marginalized}, empowerment \citep{gregor2016variational}, among many others).
Our work assumes we know that some sub-behaviours can be generically useful (like stacking or moving blocks \citep{sac-x}). We learn them using goal-conditioned policies as explained above, but we expand how these can be combined by allowing sub-behaviours to call other sub-behaviours, enabling complex hierarchies of behaviours to be discovered. We share this motivation with \citep{levy2018learning} which explores learning policies with multiple levels of hierarchy. 

Our work also relates to frameworks which decompose the value function of an MDP into combination of smaller MDPs, such as MAXQ \citep{dietterich2000hierarchical} or the factored MDP formulation \cite{guestrin2003efficient}, and more recent work which leverage natural language to achieve behaviour compositionality and generalisation \citep{language-abstraction-hrl,rl-with-sketchesr}.
As our meta-controller uses \mcts with a learned model, this also relates to advances in MBRL \citep{nagabandi2018neural,nasiriany2019planning,muzero,sekar2020planning}.
Our work extends this to the use of hierarchical programmatic skills. This closely connects to model-based HRL research area which has seen only limited attention so far \citep{silver2012compositional,nasiriany2019planning, illanes2020symbolic}.

Finally, our work capitalizes on recent advances in Neural Programmers Interpreters (NPI) \citep{npi}, in particular \anpi \citep{AlphaNPI}. \npi learns a library of program embeddings that can be recombined using a core recurrent neural network that learns to interpret arbitrary programs.
\anpi augments \npi with AlphaZero \citep{silver2016mastering,silver2017mastering} style \mcts search and RL applied to symbolic tasks such as Tower of Hanoi and sorting. We extend \anpi to challenging continuous control domains and learn a transition model over programs instead of using the environment.
Our work shares similar motivation with \citep{2017arXiv171001813X} where \npi is applied to solve different robotic tasks, but we do not require execution traces or pre-trained low-level policies.

\begin{figure*}[ht]
\centering
\includegraphics[width=0.8\textwidth]{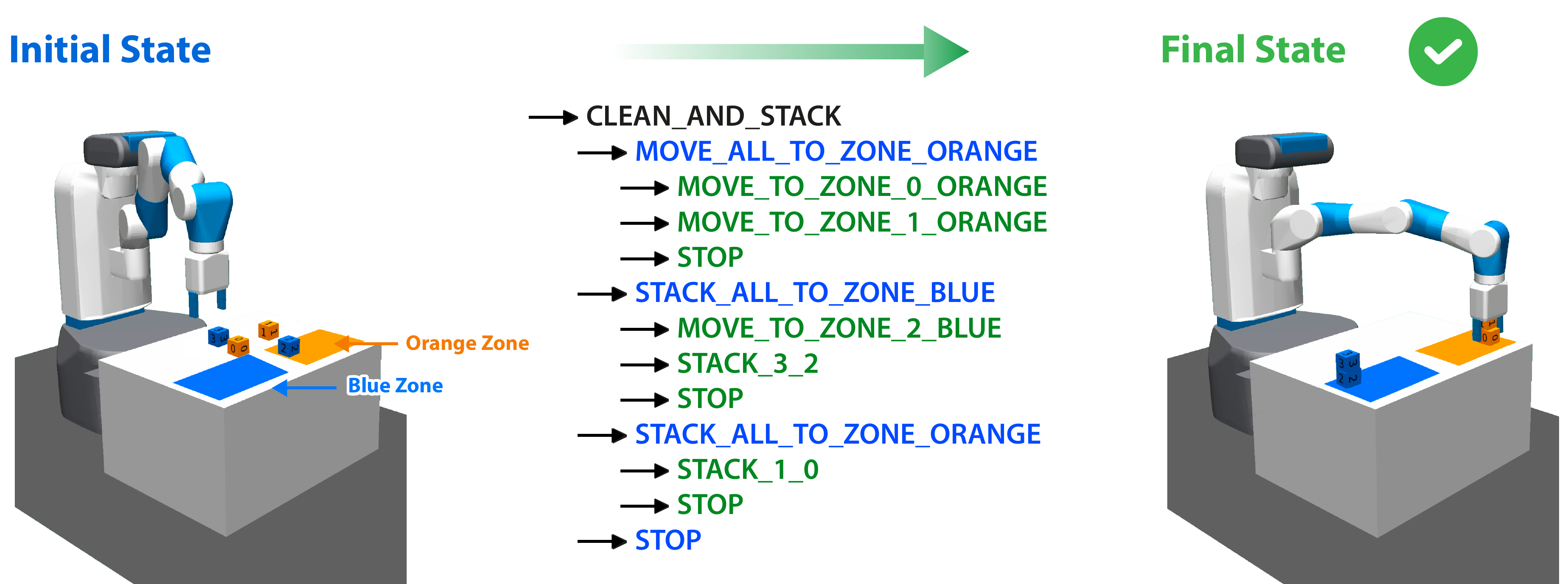}
\caption{Illustrative example of an execution trace for the {\sc Clean\_And\_Stack} program. This trace is not optimal as the program may be realised in fewer moves but it corresponds to one of the solutions found by \anx during training. \emph{Atomic} program calls are shown in green and \emph{non-atomic} program calls are shown in blue.}
\label{fig:example_trace}
\end{figure*}

\section{Problem Definition}
\label{sec:problem_definition}

In this paper, we aim to learn libraries of skills to solve a variety of tasks in continuous action domains with sparse rewards. Consider the task shown in Figure~\ref{fig:example_trace} where the agent's goal is to take the environment from its initial state where four blocks are randomly placed to the desired final state where blocks are in their corresponding coloured zones and stacked on top of each other. We formalize skills and their combinations as {\em programs}. An example programmatic trace for solving this task is shown where the sequence of programs are called to take the environment from the initial state to the final rewarding state. We specify two distinct types of programs: \emph{Atomic} programs (shown in green) are low-level goal-conditioned policies which take actions in the environment for a fixed number of steps $T$. \emph{Non-atomic} programs  (shown in blue) are a combination of atomic and/or other non-atomic programs, allowing multiple possible levels of hierarchies in behaviour. 
\begin{table}[ht]
\vskip 0.15in
\begin{center}
\begin{small}
\begin{sc}
\resizebox{0.95\textwidth}{!}{%
\begin{tabular}{lll}
\toprule
\textbf{program} & \textbf{arguments} & \textbf{description}\\
\midrule
{\sc Stack} & Two blocks id & Stack a block on another block.\\
{\sc Move\_To\_Zone} & Block id \& colour & Move a block to a colour zone.\\
\hline
{\sc Stack\_All\_Blocks}  & No arguments & Stack all blocks together in any order.\\
{\sc Stack\_All\_To\_Zone}  & colour & Stack blocks of the same colour in a zone.\\
{\sc Move\_All\_To\_Zone} & colour & Move blocks of the same colour to a zone.\\
{\sc Clean\_Table}  & No arguments & Move all blocks to their colour zone.\\
{\sc Clean\_And\_Stack} & No arguments & Stack blocks of the same colour in zones.\\
\bottomrule
\end{tabular}
}
\end{sc}
\end{small}
\end{center}
\caption{Programs library for the fetch arm environment. We obtain 20 atomic and 7 non atomic programs when considering all possible combinations when expending programs arguments. Please see Supp. Section~\ref{sec:supp_progs} for a detailed explanation of all programs and \tableautorefname{}~\ref{table:program_library_long} for all combinations when expending program arguments. 
 }\label{table:program_library}
\vskip -0.1in
\end{table} 

We base our experiments on a set of robotic tasks with continuous action space. Due to the lack of any long-horizon hierarchical multi-task benchmarks, we extended the OpenAI Gym Fetch environment \citep{openAIgym} with tasks exhibiting such requirements. We consider a target set of tasks represented by a hierarchical library of programs, see \tableautorefname{}~\ref{table:program_library}. These tasks involve controlling a robotic arm with 7-DOF to manipulate 4 coloured blocks in the environment. Tasks vary from simple block stacking to arranging all blocks into different areas depending on their colour. Initial block positions on the table and arm positions are randomized in all tasks. We consider 20 atomic programs that correspond to operating on one block at a time as well as 7 non-atomic programs that require interacting with 2 to 4 blocks. We give the full specifications of these tasks in Sup. Table~\ref{table:program_library_long}.

We consider a continuous action space $\mathcal{A}$, a continuous state space $\mathcal{S}$, an initial state distribution $\rho$ and a transition function $\mathcal{T}: \mathcal{A} \times \mathcal{S} \rightarrow \mathcal{S}$. The state vector contains the positions, rotations, linear and angular velocities of the gripper and all blocks. More precisely, a state $s_t$ has the form $s_t = [X^1_t, X^2_t, X^3_t,X^4_t, Y]$ where $X^i_t$ is the position $(x,y,z)$ of block $i$ at time step $t$ and $Y$ contains additional information about the gripper and velocities. 

More formally, we aim to learn a set of $n$ programs $p_i, \ i \in \{1, \dots, n \}$.
A program $p_i$ is defined by its pre-condition $\precond_i: \mathcal{S} \rightarrow \{ 0,1 \}$ which assesses whether the program can start and its post-condition $\postcond_i: \mathcal{S} \rightarrow \{ 0,1 \}$ which corresponds to the reward function here.
Each program is associated to an MDP $\left(\mathcal{S}, \mathcal{A}, \mathcal{T}, R_i \right)$ which can start only in states such that the pre-condition $\precond_i$ is satisfied, and where $R_i$ is a reward function that outputs $1$ when the post-condition $\psi_i$ is satisfied and 0 otherwise. 

Atomic programs are represented by a goal-conditioned neural network with continuous action space. Non-atomic programs use a modified action space: we replace the original continuous action space $\mathcal{A}$ by a discrete action space $\mathcal{A'} = \{a_1^{'}, \dots, a_n^{'},$ {\sc Stop}$\}$ where actions $a_i^{'}$ call programs $p_i$ and the {\sc Stop} action enables the current program to terminate and return the execution to its calling program. Atomic programs don't have a {\sc Stop} action, they terminate after $T$ time steps.

\section{\anx}
\label{sec:anx}

\begin{figure*}[ht]
\vskip 0.2in
\begin{center}
\centerline{\includegraphics[width=0.95\textwidth]{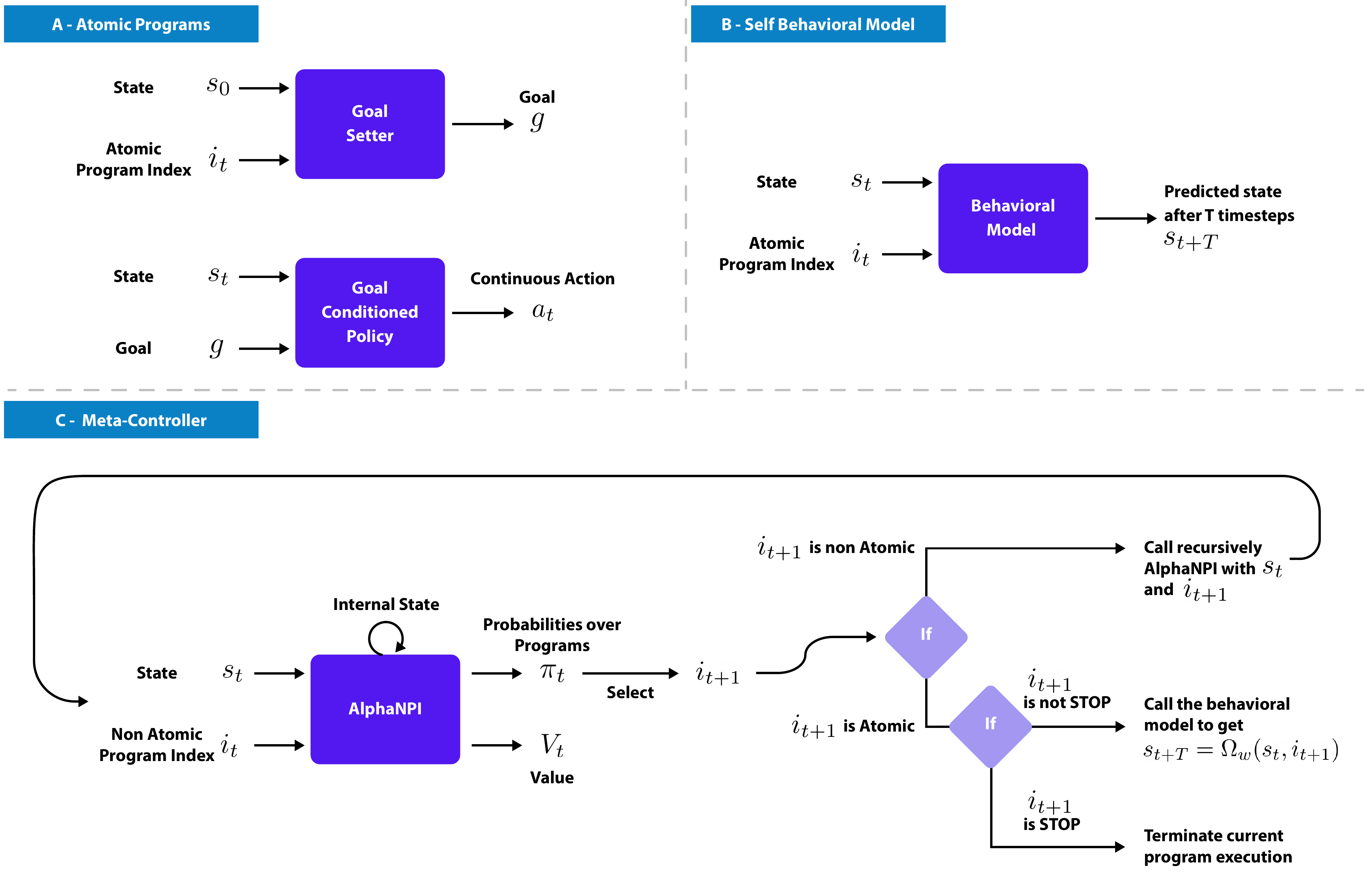}}
\caption{\anx consists of three modules: A goal setter transforms an atomic program index into a goal that is executed by a \textit{goal-conditioned policy}. A \textit{self-behavioural model} transforms an atomic program index and an environment state into a prediction of the environment state once the goal-conditioned policy has been rolled for $T$ time steps to execute this program. A \textit{meta-controller} plans through the behavioural model to execute non-atomic programs. }

\label{fig:general_method_schema}
\end{center}
\vskip -0.2in
\end{figure*}

\anx learns to solve multiple tasks by composing programs at different levels of hierarchy. Given a one-hot encoding of a program and states from the environment, the meta-controller calls either an atomic program, a non-atomic program or the {\sc Stop} action. In the next sections, we will describe how learning and inference is performed.

\subsection{Training \anx}
Learning in our system operates in three stages: first we learn the atomic programs, then we learn a transition model over their effect and finally we train the meta-controller to combine them. We provide detailed explanation of how these modules are learned below.

\subsubsection*{Learning Atomic Programs}
\label{sec:learning_atomic_programs}

Our atomic programs consist of two components: a) a \textit{goal-setter} that, given the atomic program encoding and the current environment state, generates a goal vector representing the desired position of the blocks in the environment and b) a \textit{goal-conditioned policy} that gets a “goal” as a conditioning vector, produces continuous actions and terminates after T time steps. 

To execute an atomic program $p_i$ from an initial state $s$ satisfying the pre-condition $\precond_i$, the goal-setter, $\goalgen$, computes the goal $g$: $g = \goalgen(s,i)$. We then roll the goal-conditioned policy $\atomp(. \ , g)$ for $T$ time steps to achieve the goal. In this work, the goal setter module $\Theta$ is provided which translates the atomic program specification into the corresponding goal vector indicating the desired position of the blocks which is also used to compute rewards.

We parametrize our shared goal-conditioned policy using Universal Value Function Approximators (\uvfas) \citep{uvfa}. \uvfas estimate a value function that does not only generalise over states but also goals. To accelerate training of this goal-conditioned \uvfa, we leverage the "final" goal relabeling strategy introduced in \her \citep{HER}. Past episodes of experience are relabelled retroactively with goals that are different from the goal aimed for during data collection and instead correspond to the goal achieved in the final state of the episode. The mappings from state vector to goals is simply done via extracting the blocks positions directly from the state vector $s$. To deal with continuous control, we arbitrarily use \ddpg \citep{ddpg} to train the goal-conditioned policy.

More formally, we define a goal space $\mathcal{G}$  which is a subspace of the state space $\mathcal{S}$. The goal-conditioned policy $\pi^{atomic}_{\theta}: \mathcal{S} \times \mathcal{G} \rightarrow \mathcal{A}$ takes as inputs a state $s \in \mathcal{S}$ as well as a goal $g \in \mathcal{G}$. We define the function $h: \mathcal{S} \rightarrow \mathcal{G}$ that extracts the goal $g$ from a state $s$. This policy is trained with the reward function $f_{uvfa}: \mathcal{S} \times \mathcal{G} \rightarrow \mathbbm{R}$ defined as $f_{uvfa}(s, g) = \mathbbm{1}\left( \| h(s) - g \| \leq \epsilon \right)$ where $\epsilon > 0$. The goal setter as $\goalgen: \mathcal{S} \times \{1 \dots k \} \rightarrow \mathcal{G}$ takes an initial state and an atomic program index and returns a goal such that $f_{uvfa}(s,\goalgen(s, i)) = \postcond_i(s)$. We can thus express any atomic program policy $\pi_i$ as
 \begin{equation*}
     \pi_i(.) = \pi^{atomic}_{\theta} \left( . \ , \Theta(. \ , i) \right), \ \forall i \leq k.
 \end{equation*}
We first train the policy with HER to achieve goals sampled uniformly in $\mathcal{G}$ from any state sampled from $\rho$. However, the distribution of initial states encountered by each atomic program may be very different when executing programs sequentially (as will happen when non-atomic programs are introduced).
Thus, after the initial training, we continue training the policy, but with probability $0.5$ we do not reset the environment between episodes, to approximate the real distribution of initial states which happens when atomic programs are chained together. Hence, the initial state $s$ is either sampled randomly or kept as the last state observed in the previous episode. We later show our empirical results and analysis regarding both training phases.

\begin{figure}[t]
\vskip 0.2in
\centering
\centerline{\includegraphics[width=\textwidth]{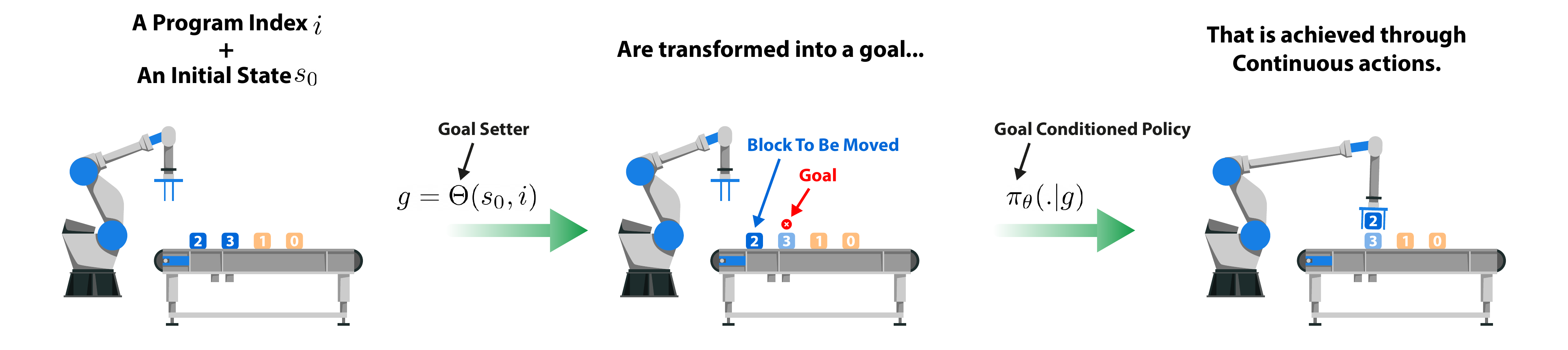}}
\caption{\anx combines a goal setter and a goal-conditioned policy to execute atomic programs. Here we see an atomic program execution for stacking block 2 on top of block 3. The goal proposed by the goal setter is shown in red which indicated the desired position for block 2. The goal-conditioned policy takes this goal as an input and receives a positive reward only if it can successfully reach this goal.}
\end{figure}
\subsubsection*{Learning Self-Behavioural Model}
\label{sec:self_behaviour_model}

After learning a set of atomic programs we learn a transition model over their effects: $\abm: \mathcal{S} \times \{ 1 \dots k \} \rightarrow \mathcal{S}$, parameterized with a neural network. As \figurename~\ref{fig:general_method_schema}-B shows, this module takes as input an initial state and an atomic program index and the output is the prediction of the environment state obtained when rolling the policy associated to this program for $T$ time steps from the initial state. We use a fully connected MLP and train it by minimizing the mean-squared error to the ground-truth final states.
This enables our model to make jumpy predictions over the effect of executing an atomic program during search, hence removing the need for any interactions with the environment during planning. 
Learning this model enables us to imagine the state of the environment following an atomic program execution and hence to avoid any further calls to atomic programs that would each have to perform many actions in the environment.

\subsubsection*{Learning the Meta-Controller}

In order to compose atomic programs together into hierarchical stacks of non-atomic programs, we use a meta-controller inspired by \anpi \citep{AlphaNPI}. The meta-controller interprets and selects the next program to execute using neural-network guided Monte Carlo Tree Search (MCTS) \citep{silver2017mastering,AlphaNPI}, conditioned on the current program index and states from the environment (see \figurename~\ref{fig:general_method_schema} for an overview of a node expansion).

We train the meta-controller using the recursive \mcts strategy introduced in \anpi \citep{AlphaNPI}: during search, if the selected action is non-atomic, we recursively build a new Monte Carlo tree for that program, using the same state $s_t$.
See Supp. Section~\ref{sec:supp_anx} for a detailed description of the search process and pseudo-code.



In \anpi \citep{AlphaNPI}, similar to AlphaZero, during the tree search, future scenarios were evaluated by leveraging the ground-truth environment, without any temporal abstraction.
Instead in this work, we do not use the environment directly during planning, but replace it by our learnt transition model over the effects of the atomic programs, the self-behavioural model described in Section~\ref{sec:self_behaviour_model}, resulting in a far more sample efficient algorithm.
Besides, \anpi used a hand-coded curriculum where the agent gradually learned from easy to hard programs.
Instead here, we randomly sample at each iteration a program to learn from, removing the need for any supervision and hand-crafting of a curriculum.
Inspired by recent work \citep{curious}, we also implemented an automated curriculum strategy based on learning progress, however we found that it does not outperform random sampling (see more detailed explanations and results in Supp. Section~\ref{sec:supp_anx}).

\subsection{Inference with \anx}
\label{sec:inference}
To infer with \anx, we compare three different inference strategies: 1) Rely only on the policy network, without planning. 2) Plan a whole trajectory using the learned transition model (i.e. self-behavioural model) and execute it fully (i.e. open-loop planning). 3) Use a receding horizon for planning inspired by Model Predictive Control (MPC) \citep{garcia1989model}. During execution, observed states can diverge from the predictions made by the self-behavioural model due to distribution shift, especially for long planning horizons, which deteriorates performance. To counter that, we do not commit to a plan for the full episode, but instead re-plan after executing any \emph{atomic} program (non-atomic programs do not trigger re-planning). The comparison between these inference methods can be found in Table~\ref{table:alphanpi_final_perf}. We also provide a detailed example in Supp. Section~\ref{sec:supp_example}.

\section{Experiments and Results}
\label{sec:experiments}

\begin{figure}[t]
    \centering
    \includegraphics[width=0.9\textwidth]{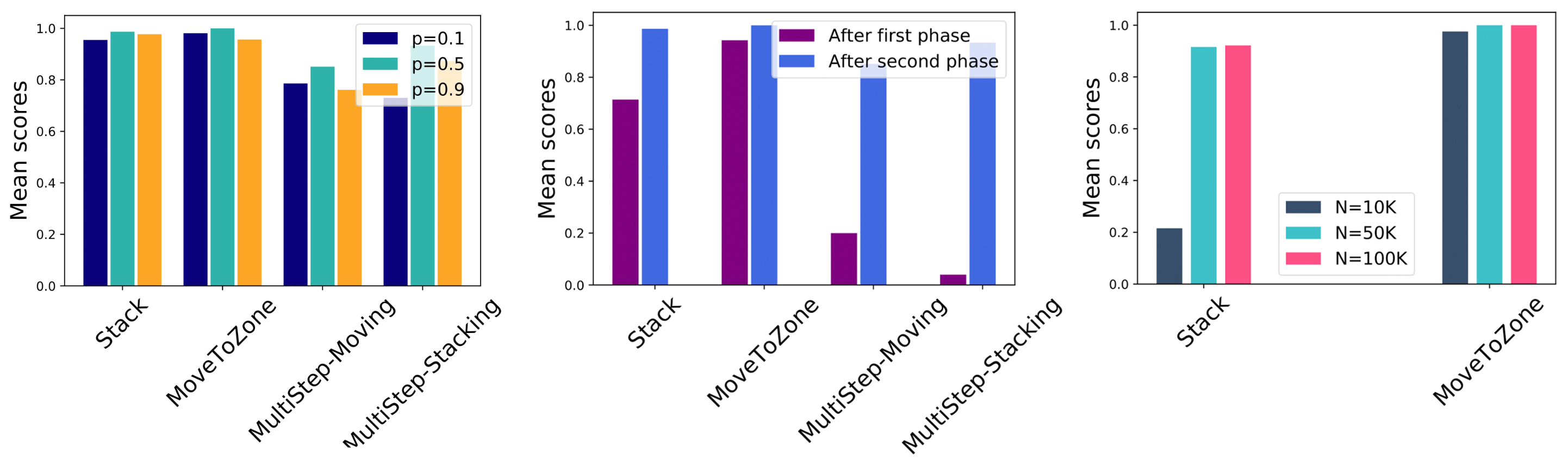}
    \caption{The goal policy $\pi^{atomic}_{\theta}$ is trained in two phases. During the first phase, goals and initial states are sampled uniformly. During the second phase, we do not reset the environment with probability $p$ between two consecutive episodes to approximate the real distribution of initial states. Once the goal policy is trained, we learn a model of its behaviour. \textbf{Left.} Comparing performance after phase 2 for different probabilities $p$. \textbf{Middle.} Comparing performance between phase 1 and phase 2. \textbf{Right.} Comparing model predictions performance when trained on different number of episodes generated with the goal policy.}\label{fig:bars_plots}
    \vspace{-10pt}
\end{figure}

We first train the atomic policies using \ddpg with \her relabelling described in Section~\ref{sec:learning_atomic_programs}. Initial block positions on the table as well as gripper position are randomized. Additional details regarding the experimental setup is provided in Supp. Section~\ref{sec:extra_experimental_details}. In the first phase of training, we train the agent for 100 epochs during uniform goal sampling. In the second phase we continue training for 150 epochs where we change the sampling distribution to allow no resets between episodes to mimic the desired setting where skills are sequentially executed.

We evaluate the agent performance on goals corresponding to single atomic programs as well as two multi-step sequential atomic program executions, \textit{MultiStep-Moving} which requires 4 consecutive calls to {\sc move\_to\_zone} and \textit{MultiStep-Stacking} which requires 3 consecutive calls to {\sc stack} (see \figurename~\ref{fig:bars_plots}). We observe that while the agent obtains decent performance on executing atomic programs in the first phase, the second phase is indeed crucial to ensure success of the sequential execution of the programs. We also observe that the reset probability affects the performance and $p=0.5$ provides the better trade-off between the asymptotic training performance and the agent ability to execute programs sequentially. More details are included in Supp. Section~\ref{sec:supp_add_exp_results}.

\begin{figure}[ht]
\vskip 0.1in
\begin{center}
\centerline{\includegraphics[width=0.55\textwidth]{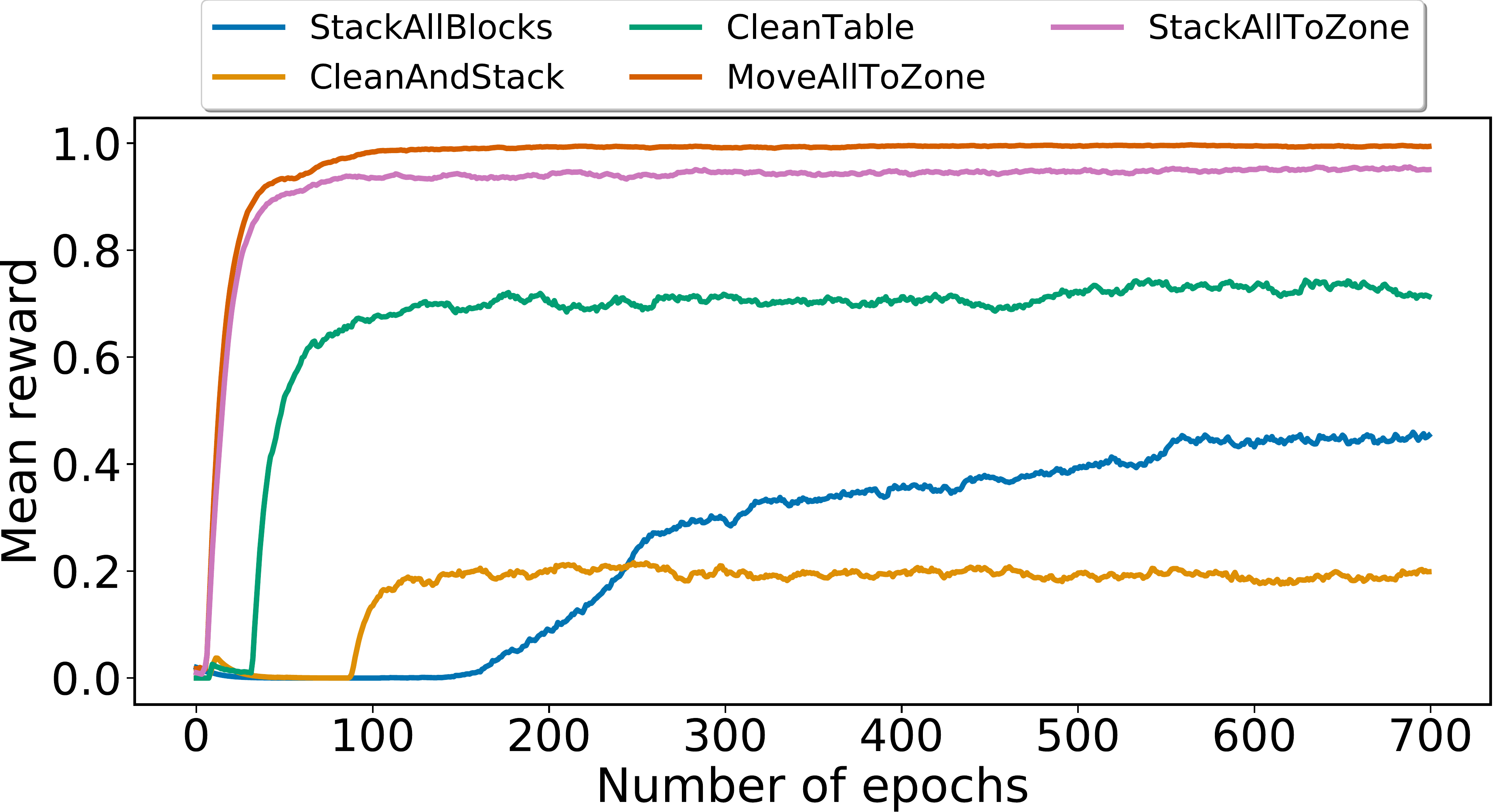}}
\caption{\anx performance evolution during training on non-atomic programs. The performance showed is the one predicted by the behavioural model and thus can differ from the one measured in the environment.}
\label{fig:metacontroller_training_evolution}
\end{center}
\vspace{-20pt}
\end{figure}

We then train the self-behavioural model to predict the effect of atomic programs, using three datasets respectively made of 10k, 50k and 100k episodes played with the goal-conditioned policy. As during the second phase of training, we did not reset the environment between two episodes with a probability $0.5$. We trained the self-behavioural model for 500 epochs on each data set. \figurename~\ref{fig:bars_plots} shows the performance aggregated across different variations of {\sc stack} and {\sc move\_to\_zone} behaviours. 

To train the meta-controller, we randomly sample from the set of non-atomic programs during training. In \figurename-\ref{fig:metacontroller_training_evolution}, we report the performance on all non-atomic tasks during the evolution of training. We also investigated the use of an automated curriculum via learning progress but observed that it does not outperform random sampling (please see Supp. Section~\ref{sec:supp_alphanpi_improvements} for details). After training for $700$ epochs, we evaluate the \anx agent on each non-atomic program in the library shown in Table~\ref{table:alphanpi_final_perf}. Our results indicate that removing planning significantly reduces performance, especially for stacking 4 blocks where removing planning results in complete failure. We observe that re-planning during inference significantly helps improve the agent's performance overall.

\begin{table}[t!]
\vskip 0.15in
\begin{center}
\begin{footnotesize}
\begin{sc}
\resizebox{0.95\textwidth}{!}{%
\begin{tabular}{l|p{10mm}p{18mm}p{16mm}|p{16mm}p{20mm}}
\toprule
\textbf{Program} & No \newline Plan  & Planning \newline & Planning \newline + {\tiny Re-planning} & Multitask \newline DDPG &Multitask \newline DDPG + HER\\
\midrule
{\sc Clean\_Table} & 0.02 & 0.54 & 0.63 & 0.0 & 0.0\\
{\sc Clean\_And\_Stack} & 0.01 & 0.25 & 0.38 & 0.0 & 0.0\\
{\sc Stack\_All\_Blocks} & 0.02 & 0.19 & 0.31 & 0.0 & 0.0\\
{\sc Stack\_All\_To\_Zone} & 0.82 & 0.92 & 0.94 & 0.0 & 0.0\\
{\sc Move\_All\_To\_Zone} & 0.68 & 0.95 & 0.95 & 0.0 & 0.0\\
\bottomrule
\end{tabular}
}
\end{sc}
\end{footnotesize}
\end{center}
\caption{ We compare the performance of \anx in different inference settings described in Section~\ref{sec:inference} as well as against 2 model-free baselines. Each program is executed 100 times with randomized environment configuration. For programs with arguments, the performance is averaged over all possible argument combinations. The full list of programs is provided in Supp. Section~\ref{sec:supp_progs}.
}\label{table:alphanpi_final_perf}
\vskip -0.1in
\end{table} 

We compare our method to two baselines to illustrate the difficulty for standard RL methods to solve tasks with sparse reward signals. First, we implemented a multitask \ddpg (\mddpg) that takes as input the environment state and a one hot encoding of a non-atomic program. At each iteration, \mddpg selects randomly a program index and plays one episode in the environment. 
Second, we implemented a \mddpg + \her agent which leverages a goal setter for non-atomic tasks described in Section~\ref{sec:learning_atomic_programs}. This non atomic programs goal setter is not available to \anx. This agent has more information than its \mddpg counterpart. Instead of receiving a program index, it receives a goal vector representing the desired end state of the blocks. As in \her, goals are relabelled during training. 
We observe that \mddpg is unable to learn any non-atomic program. It is also the case for \mddpg + \her despite having access to the additional goal representations. This shows that standard exploration mechanisms in model-free agents such as in \ddpg, where Gaussian noise is added to the actions, is very unlikely to lead to rewarding sequences and hence learning is hindered. 

\section{Conclusion}

In this paper, we proposed \anx, a novel method for constructing a hierarchy of abstract actions in a rich object manipulation domain with sparse rewards and long horizons. Several ingredients proved critical in the design of \anx. First, by learning a self-behavioural model, we can leverage the power of recursive AlphaZero-style look-ahead planning across multiple levels of hierarchy, without ever interacting with the real environment. Second, planning with a receding horizon at inference time resulted in more robust performance. We also observed that our approach does not require a carefully designed curriculum commonly used in NPI so random task sampling can be simply used instead. Experimental results demonstrated that \anx, using an abstract imagination-based reasoning, can simultaneously solve multiple complex tasks involving dexterous object manipulation beyond the reach of model-free methods.

A limitation of our work is that similar to \anpi, we pre-specified a hierarchical library of the programs to be learned. While it enables human interpretability, this requirement is still quite strong. Thus, a natural next step would be to extend our algorithm to discover these programmatic abstractions during training so that our agent can specify and learn new skills hierarchically in a fully unsupervised manner.

\section*{Broader Impact}

Our paper presents a sample efficient technique for learning in sparse-reward temporally extended settings. It does not require any human demonstrations and can learn purely from sparse reward signals. Moreover, after learning low-level skills, it can learn to combine them to solve challenging new tasks without any requirement to interact with the environment. We believe this has a positive impact in making reinforcement learning techniques more accessible and applicable in settings where interacting with the environment is costly or even dangerous such as robotics. Due to the compositional nature of our method, the interpretability of the agent's policy is improved as high-level programs explicitly indicate the intention of the agent over multiple steps of behaviour. This participates in the effort of building more interpretable and explainable reinforcement learning agents in general. 
Furthermore, by releasing our code and environments we believe that we help efforts in reproducible science and allow the wider community to build upon and extend our work in the future.

\section*{Acknowledgements}

Work by Nicolas Perrin was partially supported by the French National Research Agency (ANR), Project ANR-18-CE33-0005 HUSKI.

\bibliography{ms}
\bibliographystyle{plainnat}

\clearpage
\newpage
\appendix
\section*{Supplementary material}

\section{Programs library}
\label{sec:supp_progs}

Our program library includes 20 atomic programs and 7 non-atomic ones. The full list of these programs can be viewed in Table~\ref{table:program_library_long}. In the \anpi paper, the program library contained only five atomic programs and 3 non-atomic programs. Thus, the branching factor in the tree search in \anx is on average much greater than in the \anpi paper. Furthermore, in this work, these atomic programs are learned and therefore might not always execute as expected while in \anpi the five atomic programs are hard-coded in the environment and thus execute successfully anytime they are called.

While in this study we removed the need for a hierarchy of programs during learning, we still defined programs levels for two purposes: (i) to control the dependencies between programs, programs of lower levels cannot call programs of higher levels; (ii) to facilitate the tree search by relying on the level balancing term $L$ introduced in the original \anpi P-UCT criterion. In this context, we defined \textsc{Clean\_Table} and \textsc{Clean\_And\_Stack} as level 2 programs, \textsc{Stack\_All\_Blocks}, \textsc{Move\_All\_To\_Zone} and \textsc{Move\_All\_To\_Zone} as level 1 programs. The atomic programs are defined as level 0 programs. Interestingly, the natural order learned during training (when sampling random program indices to learn from) matches this hierarchy, see \figureautorefname{}~\ref{fig:metacontroller_training_evolution_comparison}.

\begin{table}[h]
\vskip 0.15in
\begin{center}
\begin{small}
\begin{sc}
\begin{tabular}{lll}
\toprule
\textbf{program} & \textbf{description}\\
\midrule
{\sc Stack\_0\_1} & Stack block number 0 on block number 1.\\
{\sc Stack\_0\_2} & Stack block number 0 on block number 2.\\
{\sc Stack\_0\_3} & Stack block number 0 on block number 3.\\
{\sc Stack\_1\_0} & Stack block number 1 on block number 0.\\
{\sc Stack\_1\_2} & Stack block number 1 on block number 2.\\
{\sc Stack\_1\_3} & Stack block number 1 on block number 3.\\
{\sc Stack\_2\_0} & Stack block number 2 on block number 0.\\
{\sc Stack\_2\_1} & Stack block number 2 on block number 1.\\
{\sc Stack\_2\_3} & Stack block number 2 on block number 3.\\
{\sc Stack\_3\_0} & Stack block number 3 on block number 0.\\
{\sc Stack\_3\_1} & Stack block number 3 on block number 1.\\
{\sc Stack\_3\_2} & Stack block number 3 on block number 2.\\
{\sc Move\_To\_Zone\_0\_ORANGE} & Move block number 0 to orange zone.\\
{\sc Move\_To\_Zone\_1\_ORANGE} & Move block number 1 to orange zone.\\
{\sc Move\_To\_Zone\_2\_ORANGE} & Move block number 2 to orange zone.\\
{\sc Move\_To\_Zone\_3\_ORANGE} & Move block number 3 to orange zone.\\
{\sc Move\_To\_Zone\_0\_BLUE} & Move block number 0 to blue zone.\\
{\sc Move\_To\_Zone\_1\_BLUE} & Move block number 1 to blue zone.\\
{\sc Move\_To\_Zone\_2\_BLUE} & Move block number 2 to blue zone.\\
{\sc Move\_To\_Zone\_3\_BLUE} & Move block number 3 to blue zone.\\
\hline
{\sc Stack\_All\_To\_Zone\_ORANGE}  & Stack orange blocks in the orange zone.\\
{\sc Stack\_All\_To\_Zone\_BLUE}  & Stack blue blocks in the blue zone.\\
{\sc Move\_All\_To\_Zone\_ORANGE} & Move orange blocks to the orange zone.\\
{\sc Move\_All\_To\_Zone\_BLUE} & Move blue blocks to the orange zone.\\
{\sc Stack\_All\_Blocks}  & Stack all blocks together in any order.\\
{\sc Clean\_Table}  & Move all blocks to their colour zone.\\
{\sc Clean\_And\_Stack} & Stack blocks of the same colour in zones.\\
\bottomrule
\end{tabular}
\end{sc}
\end{small}
\end{center}
\caption{Programs library for the fetch arm environment. We show the flat list of all program possibilities by expending their arguments. }\label{table:program_library_long}
\vskip -0.1in
\end{table}

\newpage
\section{Details of the \anx method}
\label{sec:supp_anx}
\subsection{Table of symbols}
\label{sec:supp_symbols}
Here we provide the list of symbols used in our method section:

\begin{table}[h!]
    \centering
    \begin{tabular}{|l|l|l|}
  \hline
    \textbf{Name} & \textbf{Symbol} & \textbf{Notes} \\
  \hline
  Action space & $\mathcal{A}$ & $\mathcal{A}$ is continuous\\
  State space & $\mathcal{S}$ & $\mathcal{S}$ is continuous\\
  Goal space & $\mathcal{G}$ & $\mathcal{G}$ is a sub-space of $\mathcal{S}$\\
  Initial state distribution & $\rho$ & \\
  Reward obtained at time $t$ & $r_t$ &  \\
  Discount factor & $\gamma$ & \\
  Program & $p_i, \ i \in \{1 \dots n\}$ &  If $i \leq k$ then $p_i$ is atomic\\
  Program index selected at time $t$ & $i_t$ & \\
  Environment state & $s_t$ & \\
  Program $p_i$ pre-condition & $\precond_i: \mathcal{S} \rightarrow \{0,1\}$ & \\
  Program $p_i$ post-condition & $\postcond_i: \mathcal{S} \rightarrow \{0,1\}$ & \\
  \anx policy & $\pi: \mathcal{S} \times \{1 \dots n\} \rightarrow \mathcal{A}$ & \\
  Program $p_i$ policy & $\pi_i$ & $\forall s \in \mathcal{S} \mid \precond(s)=1, \ \pi_i(s) = \pi(s, i)$\\
  Goal setter & $\goalgen: \mathcal{S} \times \{1 \dots k \} \rightarrow \mathcal{G}$ & \\
  \her mapping function & $h: \mathcal{S} \rightarrow \mathcal{G}$ & extract a goal from a state vector\\
  Goal conditioned policy & $\atomp: \mathcal{S} \times \mathcal{G} \rightarrow \mathcal{A}$ & $\pi_i(.) = \atomp(., \goalgen(.,i))$ \\
  Self-behavioural model & $\abm: \mathcal{S} \times \{1 \dots k \} \rightarrow \mathcal{S}$ & \\
  
  \hline
  \end{tabular}
  \vspace{1mm}
    \caption{\anx table of symbols}
    \label{tab:symbols}
\end{table}

\subsection{AlphaNPI}
\label{sec:supp_alphanpi}

Here we provide some further details on the \anpi method, which we use and extend for our meta-controller. The \anpi agent uses a recursion augmented Monte Carlo Tree Search (MCTS) algorithm to learn libraries of hierarchical programs from sparse reward signals. The tree search is guided by an actor-critic network inspired from the Neural Programmers-Interpreter (NPI) architecture. A stack is used to handle programs as in a standard program execution: when a non-atomic program calls another program, the current NPI network's hidden state is saved in a stack and the next program execution starts. When it terminates, the execution goes back to the previous program and the network gets back its previous hidden state. The mechanism is also applied to MCTS itself: when the current search decides to execute another program, the current tree is saved along with the network's hidden state in the stack and a new search starts to execute the desired program.

\begin{figure*}[ht]
\centering
\includegraphics[width=0.8\textwidth]{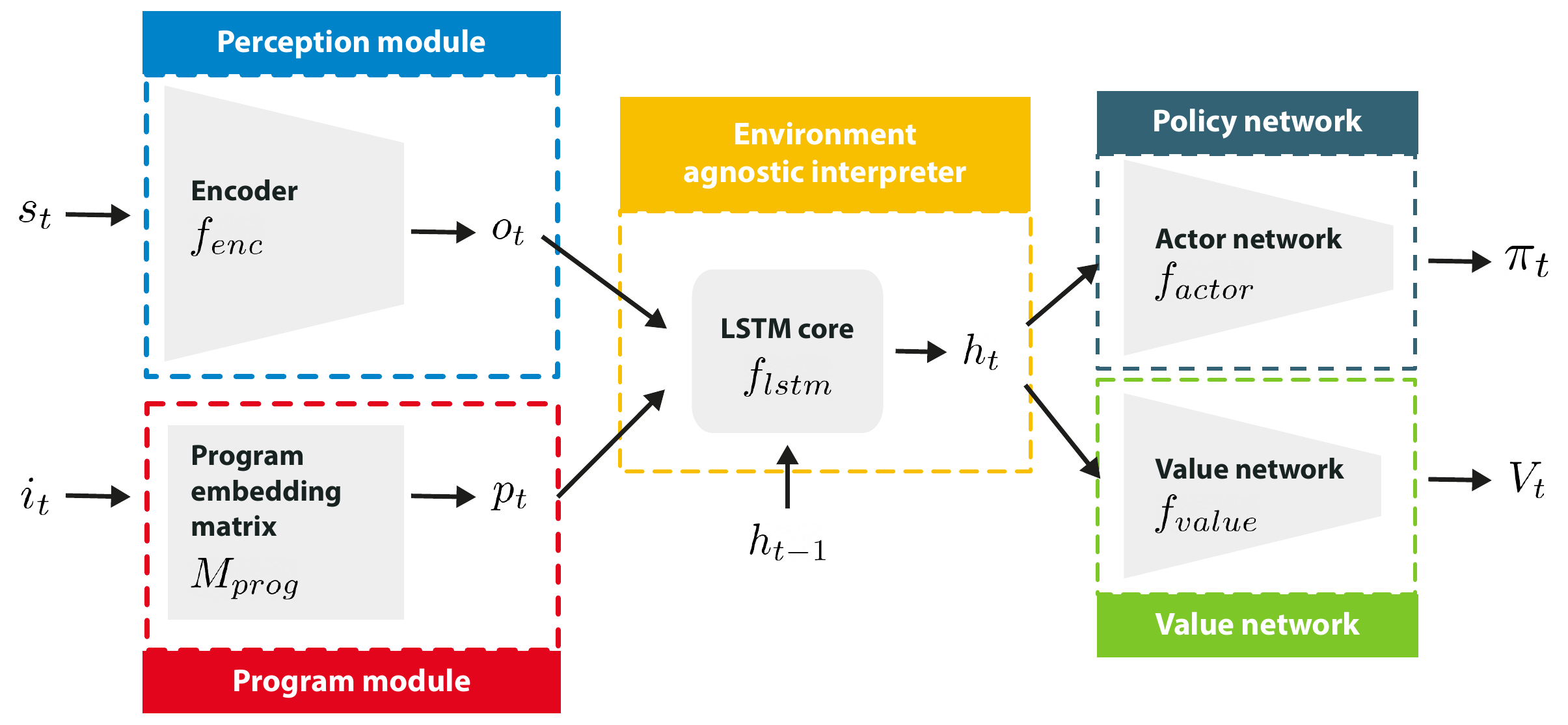}
\caption{The \anpi architecture, slightly modified from \cite{AlphaNPI}.}
\label{fig:anpi_archi}
\end{figure*}

The neural network takes as input an environment state $s_t$ and a program index $i_t$ and returns a vector of probabilities over the next programs to call, $\pi_t$, as well as a prediction for the value $V_t$.
The network is composed of five modules. An encoder $f_{enc}$ encodes the environment state into a vector $o_t = f_{enc}(s_t)$ and a program embedding matrix $M_{prog}$ contains learnable embedding for each non-atomic programs. The $i$-th row of the matrix contains the embedding $p$ of the program referred by the index $i$ such that $p_t = M_{prog}(i_t)$. An LSTM core $f_{lstm}$ takes the encoded state and the program embedding as input and returns its hidden state $h_t$. Finally, a policy head and a value head take the hidden state as input and return respectively $\pi_t$ and $V_t$, see \figureautorefname{}~\ref{fig:anpi_archi}.

The guided MCTS process is used to generate data to train the AlphaNPI neural network. A search returns a sequence of transitions $(s_t, i_t, \pi^{mcts}_t, r)$ where $\pi^{mcts}_t$ corresponds to the tree policy at time $t$ and $r$ is the final episode reward. The tree policy is computed from the tree nodes visit counts. These transitions are stored in a replay buffer. At training time, batches of trajectories are sampled and the network is trained to minimize the loss

\begin{equation}
    \label{eq:loss_fct}
    \ell = \sum_{\text{batch}} - \underbrace{\left( \pi^{mcts} \right)^T \log{ \pi }}_{\ell_{\text{policy}}} + \underbrace{(V - r)^2}_{\ell_{\text{value}}}.
\end{equation}

Note that in \anx, in contrast with the original \anpi work, proper back-propagation through time (BPTT) over batches of trajectories is performed instead of gradient descent over batches of transitions, resulting in a more stable updates.

\anpi can explore and exploit. In exploration mode, many simulations per node are performed, Dirichlet noise is added to the priors to help exploration and actions are chosen by sampling the tree policy vectors. In exploitation mode, significantly less simulations are used, no noise is added and actions are chosen by taking the argmax of the tree policy vectors. The tree is used in exploration mode to generate data for training and in exploitation mode at inference time.


\subsection{Improvements over AlphaNPI}
\label{sec:supp_alphanpi_improvements}
\paragraph{Distributed training and BPTT}
We improved \anpi training speed and stability in distributing the algorithm. We use 10 actors in parallel. Each actor has a copy of the neural networks weights. It uses it to guide its tree searches to collect experiences. In each epoch, the experience collected by all the actors is sent to a centralized replay buffer. A learner, that has a copy of the network weights, samples batches in the buffer, computes the losses and performs gradient descent to update the network weights. When the learner is done, it sends the new weights to all the actors. We use the MPI (Message Passing Interface) paradigm through its python package MPI4py to implement the processes in parallel. Each actor uses 1 CPU.

Another improvement over the standard \anpi is leveraging back-propagation through time (BPTT). The \anpi agent uses an \lstm. During tree search, the \lstm states are stored inside the tree nodes. However, in the original \anpi, transitions are stored independently and gradient descents are performed on batches of uncorrelated transitions. While it worked on the examples presented in the original paper, we found that implementing BPTT improves training stability.

\paragraph{Curriculum Learning}
At each training iteration, the agent selects the programs to train on. In the original \anpi paper, the agents select programs following a hard-coded curriculum based on program complexity. In this work, we select programs randomly instead and thus remove the need for extra supervision. We also implemented an automated curriculum learning paradigm based on learning progress signal \cite{curious} and observed that it does not improve over random sampling. We compare both strategies in \figureautorefname{}~\ref{fig:metacontroller_training_evolution}. We detail below the learning progress based curriculum we used as a comparison.

The curriculum based on the program levels from the original \anpi might be replaced by a curriculum based on learning progress. The agent focuses on programs for which its learning progress is maximum. This strategy requires less hyper-parameters and enables the agent to discover the same hierarchy implicitly. The learning progress for a program $i$ is defined as the derivative over time of the agent performance on this program. At the beginning of each training iteration $t$, the agent attempts every non atomic program $l$ times in exploitation mode. We note $C^i_t$ the average performance, computed as mean reward over the $l$ episodes, on program $i$ at iteration $t$. Rewards are still assumed to be binary: 0 or 1. We compute the learning progress for program $i$ as

\begin{equation}
    LP^i_{t} = \left| C^i_{t} - C^{i}_{t-1} \right|,
\end{equation}

where the absolute derivative over time of the agent performance is approximated as the first order. Finally, we compute the probability $p_t^i$ of choosing to train on the program $i$ at iteration $t$ as

\begin{equation}
    p^i_{t} = \epsilon * \frac{1}{M} + \left( 1 - \epsilon \right) * \frac{LP^i_{t}}{\sum_{j} LP^j_{t}},
\end{equation}

where $\epsilon \in [0,1]$ is an hyperparameter and $M$ is the total number of non atomic programs. The term $\epsilon * \frac{1}{M}$ is used to balance exploration and exploitation, it ensures that the agent tries all programs with a non-zero probability.

\newpage

\subsection{Pseudo-codes}
\label{sec:supp_pseudo_codes}
\subsection*{Goal policy training}
\begin{algorithm}[!ht]
\SetAlgorithmName{Algorithm}{}
\KwData{}
{\small

\For{num\_epoch}{

    \For{num\_cycle}{
    
        \For{num\_ep\_per\_cycle}{
        
            Reset the environment in a state $s_0 \sim \rho$
            
            Sample uniformly in the goal space a goal $g \sim \mathcal{G}$
            
            Play one episode with the goal policy $\atomp(. \ | g)$ from $s_0$
            
            Store the trajectory transitions in a replay buffer
        }
        
        \For{num\_sgd\_per\_cycle}{
        
            Sample a batch of transitions in the replay buffer
            
            Resample according to \her strategy
            
            Compute the gradient on this batch
            
            Update weights $\theta$ with the gradient averaged over all actors
        }
    }
    
}
}
\caption{Goal policy training: First phase}
\label{alg:goal_policy_first}
\end{algorithm}
\begin{algorithm}[!ht]
\SetAlgorithmName{Algorithm}{}
\KwData{}
{\small

\For{num\_epoch}{

    \For{num\_cycle}{
    
        \For{num\_ep\_per\_cycle}{
        
            Choose uniformly a motion program $p_i$
            
            With probability 0.5 reset the environment
            
            Otherwise, sample a state $s_0$ such that $\precond_i(s_0) = 1$ in the state buffer and reset to this state
            
            Compute with the goal setter a goal $g = \goalgen(s_0, i)$ corresponding to this program
            
            Play one episode with the goal policy $\atomp(. \ | g)$ from $s_0$
            
            Store the trajectory in a replay buffer
            
            Store the final state in the state buffer
        }
        
        \For{num\_sgd\_per\_cycle}{
        
            Sample a batch of transitions in the replay buffer
            
            Resample according to \her strategy
            
            Compute the gradient on this batch
            
            Update weights $\theta$ with the gradient averaged over all actors
        }
    }
    
}
}
\caption{Goal policy training: Second phase}
\label{alg:goal_policy_second}
\end{algorithm}

\newpage
\subsection*{Self-behavioural model training}
\begin{algorithm}[!ht]
\SetAlgorithmName{Algorithm}{}
\KwData{}
{\small
\textbf{Data generation}

\For{num\_episode}{

    Choose uniformly a motion program $p_i$
    
    With probability 0.5 reset the environment
    
    Otherwise, sample a state $s_0$, such that $\precond_i(s_0) = 1$, in the state buffer and reset to this state
    
    Compute with the goal setter a goal $g=\goalgen(s_0,i)$
    
    Play one episode with the goal policy $\atomp(. \ |, g)$ from $s_0$
    
    Store the initial state $s_0$, the final state $s_f$ and the program index $i$ in a dataset
    
}
\texttt{\\}
\textbf{Training}

\For{num\_epochs}{
    \For{num\_sgd\_epoch}{
        Sample a batch of $\left( s_0, s_f, i \right)$ tuples
        
        Update $w$ to minimize $l = \sum\limits_{batch} \left( \abm(s_0, i) - s_f \right)^2$
    }
}
}
\caption{Learning the behavioral model of the APEM}
\label{alg:apem_model_training}
\end{algorithm}

\subsection*{AlphaNPI training}

\begin{algorithm}[!ht]
\SetAlgorithmName{Algorithm}{}
\KwData{}
{\small

\For{num\_epoch}{

    \For{num\_task\_per\_epoch}{
    
    Sample program $p_i$ according to the curriculum strategy
    
        \For{num\_ep\_per\_task}{
        
            Sample an initial state $s_0 \in \mathcal{S}$ until $\precond_{i}(s_0) = 1$
            
            Run an AlphaNPI search using the self behavioral model to obtain a final state $s_f$
            
            Compute the reward $r = \postcond_i(s_f)$
            
            Store the trajectory in a replay buffer
        }
    }
    
    \For{num\_sgd\_per\_epoch}{
        
        Sample a minibatch of transitions in the replay buffer
        
        Train the AlphaNPI neural net with it
    }
    
    \For{every program}{
        
        Play $n\_val$ episodes with the tree in exploitation mode
        
        Record the averaged score
    }
}
}
\caption{AlphaNPI training}
\label{alg:alphanpi_training}
\end{algorithm}
\newpage
\subsection*{AlphaNPI inference}

\begin{algorithm}[h]
\SetAlgorithmName{Algorithm}{}
\KwData{}
{\small
\textbf{Input}: program index $i$ and an initial state $s_0$

\While{$True$}{
    
    Run $n_{simu}$ with AlphaNPI relying on the self behavioral model from current state $s_{t}$
    
    Compute tree policy $\pi^{mcts}$ with visit counts
    
    Select the next program to call $p_i$
    
    \eIf{$i \leq k$}{
        Roll $\atomp(. \ | \goalgen(s_t, i))$ for $T$ timesteps to get new state $s_{t+T}$
    }{
        \eIf{$a == $ STOP}{
         Stop the search\;
       }{
         Run a new search recursively to execute the program
       }
   
    }
}    
}
\caption{AlphaNPI execution with MPC during inference}
\label{alg:alphanpi_testing}
\end{algorithm}

\clearpage


\section{Example execution of the programs}
\label{sec:supp_example}

Let us imagine we want to run the atomic program {\sc Stack\_1\_2}. The state space is $\mathcal{S}=\reals^{70}$ and the goal space is $\mathcal{G} = \reals^{12}$. A state $s_t$ contains all positions, angles, velocities and angular velocities of the blocks and of the robot articulations  at time $t$. A goal $g$ is the concatenation of desired final positions $(x,y,z)$ for each block. Let us consider atomic program {\sc Stack\_1\_2} has index 0. In this case, we compute the goal corresponding to this program as $g = \goalgen(s_0, 0) = (x^2_0, y^2_0, z^2_0 + 2d, \ x^2_0, y^2_0, z^2_0, \ x^3_0, y^3_0, z^3_0, \ x^4_0, y^4_0, z^4_0)$ where $d$ is the block radius. This goal corresponds to positions where block 1 is stacked on block 2 and where other blocks have not been moved. The pre-condition of this program verifies in state $s_0$ that no other block is already on top of block $2$ so that stacking is possible. If the pre-condition is verified, the goal policy is rolled for $T=50$ time steps to execute this goal. When the goal policy terminates, the program post-condition is called on the final state $s_T$ and verifies for all blocks that their final position is in a sphere of radius $\epsilon$ around their expected final position described by goal $g$. The value we use for $\epsilon$ is the standard value used in \her \citep{HER}.

We train the goal-conditioned policy $\atomp$ so that it can reach goals $g$ that corresponds to the atomic programs for any initial position, as long as the initial position verifies the atomic program pre-condition. Once trained, we learn its self behavioural model $\abm$. This model takes an environment state and an atomic program index and predicts the environment state obtained when the goal policy has been rolled for $T$ timesteps conditioned on goal and initial environment state. Thus, this model performs jumpy predictions, it does not capture instantaneous effects but rather global effects such as \textit{block $0$ has been moved, block $1$ has remained at its initial position}.

When the model has been trained, we use it to learn non-atomic programs with our extended \anpi. Let us imagine that we want to execute the {\sc Clean\_And\_Stack} program. This program is expected to stack, in any order, both orange blocks in the orange zone and both blue blocks in the blue zone. Its pre-condition is always true since it can be called from any initial position. Its post-condition looks at a state $s$ and returns $1$ if it finds an orange block with its center of gravity in the orange zone and with an orange block stacked in top of it. It verifies the same for the blue blocks. The post-condition returns $0$ otherwise. The stacking test is performed as explained above looking at a sphere of radius $\epsilon$ around an ideal position.

A possible execution trace for this program is:


\begin{figure*}[ht]
\centering
\includegraphics[width=0.8\textwidth]{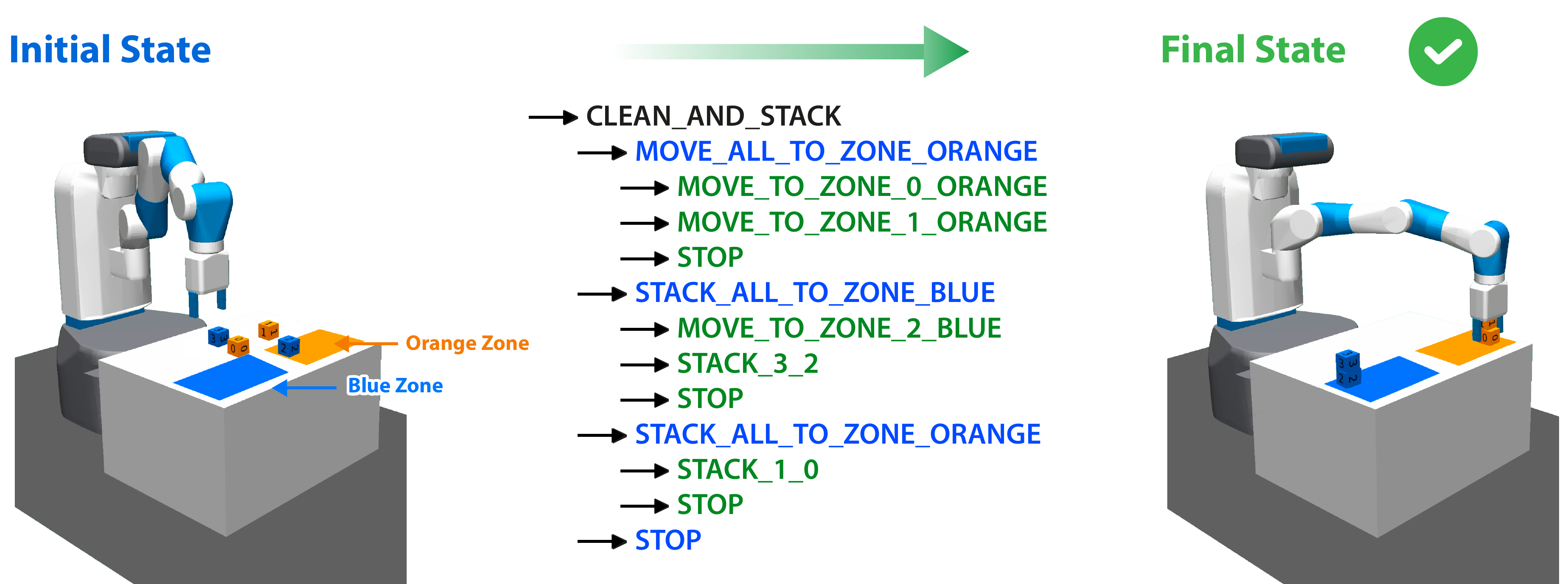}
\caption{Illustrative example of an execution trace for the {\sc Clean\_And\_Stack} program. This trace is not optimal as the program may be realised in fewer moves but it corresponds to one of the solutions found by \anx during training. \emph{Atomic} program calls are shown in green and \emph{non-atomic} program calls are shown in blue.}
\label{fig:example_trace_app}
\end{figure*}

In this trace, the non-atomic programs are shown in blue and the atomic programs are shown in green.  Non-atomic program execution happens as in the original \anpi. The execution of an atomic program happens as described before: when the atomic program is called, the current environment state is transformed into a goal vector that is realised by the goal-conditioned policy. In the trace above, the goal policy has been called $5$ times with different goal vectors.

\section{Experimental setting}
\label{sec:extra_experimental_details}

\subsection{Experiment setup}
\begin{figure}[ht]
\vskip 0.2in
\begin{center}
\centerline{\includegraphics[width=0.7\textwidth]{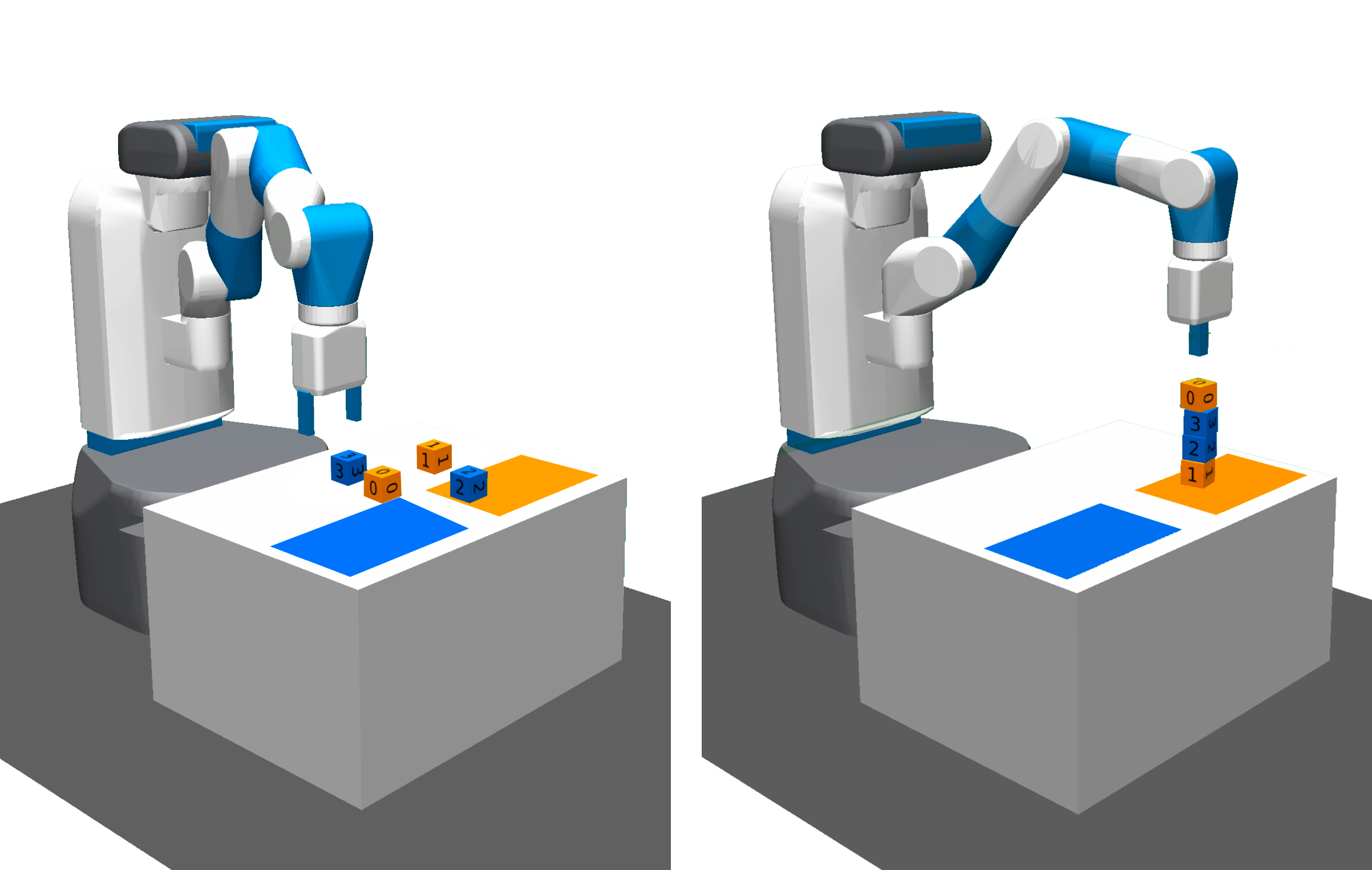}}
\caption{\textbf{Our multi-task fetch arm environment.} The agent first learns to manipulate all blocks on the table. Then it learns to perform a large set of manipulation tasks. These tasks require abstraction as well as precision. The agent learns to move blocks on zones depending of the block color. It learns to stack all blocks together and to stack blocks of the same color in the corresponding zone. }
\label{fetch_environment}
\end{center}
\vskip -0.2in
\end{figure}

We base our experiments on a set of robotic tasks with continuous action space. Due to the lack of any
long-horizon hierarchical multi-task benchmarks, we extended the OpenAI Gym Fetch environment
\cite{openAIgym} with tasks exhibiting such requirements. We reused the core of the {\sc Pick And Place} environment and added two colored zones as well as four colored blocks that are indexed. We did not modify the environment physics. The action space is $\mathcal{A} = [-1,1]^4$, it corresponds to continuous command $dx,dy,dz,d\alpha$ to the arm gripper. In this setting, the observation space is the same as the state space and contains all blocks and robot joint positions together with their linear and angular velocities. It is of dimension $70$, with blocks ordered in the observations removing the need to provide their colour or index. More precisely, a state $s_t$ has the form $s_t = [X^1_t, X^2_t, X^3_t,X^4_t, Y]$ where $X^i_t$ is the position $(x,y,z)$ of block $i$ at time step $t$ and $Y$ contains additional information about the gripper and velocities. When executing atomic programs, we assume that we always aim to move the block which is at the first position in the observation vector. Then, to move any block $j$, we apply a circular permutation to the observation vector such that block $j$ arrives at the first position in the observation. 

We consider two different initial state distributions $\rho$ in the environment. During goal-conditioned policy training, the block at the first position in the observation vector has a $0.2$ probability to start in the gripper, a $0.4$ probability to start under the gripper and a $0.4$ probability to start elsewhere uniformly on the table. All other blocks may start anywhere on the table. Once the first block is positioned, we place the other blocks in a random order. When a block is placed, it has a $0.4$ probability to be placed on top of another block and $0.6$ to be placed on the table. In this case, its position is sampled uniformly on the table until there is no collision. During training, we anneal these probabilities. By the end of training all blocks are always initialised at random positions on the table.

\subsection{Computational resources}
We ran all experiments on a 30 CPU cores. We used 28 CPUs to train the goal-conditioned policy and 10 CPU cores to train the meta-controller. We trained the goal-conditioned policy for 2 days (counting both phases) and the meta-controller for also 2 days. 

\newpage
\subsection{Hyper-parameters}

We provide all hyper-parameters used in our experiments for our method together with baselines.

\begin{table}[h!]
\centering
\begin{tabular}{|l|l|r|}
\hline
\textbf{Notation} & \textbf{Description} & \textbf{Value} \\
\hline
\noalign{\vskip 2mm}
\hline
\multicolumn{3}{|l|}{\textbf{Atomic Programs}}\\
\hline
DDPG actor & hidden layers size & 256/256\\
DDPG critic & hidden layers size & 256/256\\
$\gamma$ & discount factor & 0.99\\
{\it batch size} & batch size (number of transitions) & 256\\
$lr_{actor}$ & actor learning rate & $0.001$\\
$lr_{critic}$ & critic learning rate & $0.001$\\
{\it replay\_k} & ratio to resample transitions in HER & 4\\
$n_{actors}$ & number of actors used in parallel & 28\\
$n_{cycles\_per\_epoch}$ & number of cycles per epoch & 40\\
$n_{batches}$ & number of updates per cycle & 40\\
buffer size & replay buffer size (number of transitions) & $1000000$\\
$n_{epochs\_phase\_1}$ & number of epochs for phase 1 & 100\\ 
$n_{epochs\_phase\_2}$ & number of epochs for phase 2 & 150\\ 
\hline
\noalign{\vskip 4mm}
\hline
\multicolumn{3}{|l|}{\textbf{Behavioural Model}}\\
\hline
Model layers & hidden layers size & 512/512 \\
Dataset size & number of episodes collected to create dataset & 50000\\
$n_{epochs}$ & number of epochs & 500\\
\hline
\noalign{\vskip 4mm}
\hline
\multicolumn{3}{|l|}{\textbf{AlphaNPI}}\\
\hline
Observation Module & hidden layer size & 128\\
$P$ & program embedding dimension & 256\\
$H$ & \lstm hidden state dimension & 128\\
$S$ & observation encoding dimension & 128\\
$\gamma$ & discount factor to penalize long traces reward & 0.97\\
$n_{simu}$ & number of simulations in the tree in exploration mode & 100 \\
$n_{simu-exploit}$ & number of simulations in the tree in exploitation mode & 5\\
$n_{updates\_per\_episode}$ & number of gradient descent per episode played & 2\\
$n_{ep}$ & number of episodes at each iteration & 20\\
$n_{val}$ & number of episodes for validation & 10\\
$n_{iteration}$ & number of iterations & 700\\
$c_{\text{level}}$ & coefficient to encourage choice of higher level programs & 3.0\\
$n_{actors}$ & number of actors used in parallel & 10\\
$c_{\text{puct}}$ & coefficient to balance exploration/exploitation in \mcts & 0.5\\
{\it batch size} & batch size (number of trajectories) & 16\\
$n_{buf}$ & maximum size of buffer memory (nb of full trajectories) & 100\\
$p_{buf}$ & probability to draw positive reward experience in buffer & 0.5\\
$lr$ & learning rate & $0.0001$\\
$\tau$ & tree policies temperature coefficient & 1.3\\
$\epsilon_d$ & AlphaZero Dirichlet noise fraction & 0.25\\
$\alpha_d$ & AlphaZero Dirichlet distribution parameter & 0.03\\
\hline
\end{tabular}
\vspace{1mm}
\caption{Hyperparameters}
\label{table:hyperparameters_table}
\end{table}

\newpage

\section{Additional Results}
\label{sec:supp_add_exp_results}

\subsection{Results of Goal Policy Training}

\begin{figure}[ht]
\centering
\begin{subfigure}
  \centering
  \includegraphics[width=0.49\linewidth]{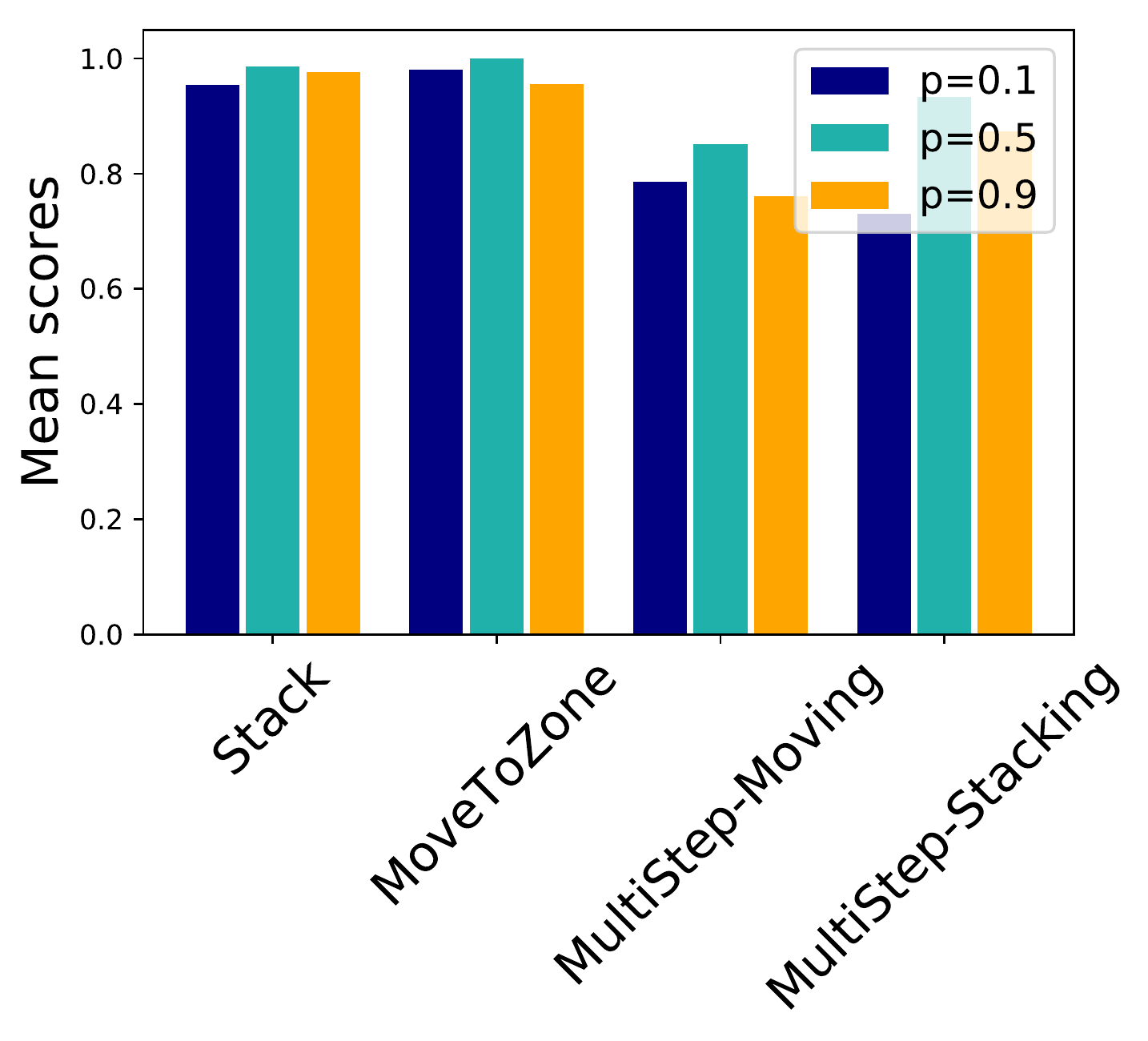}
  \label{fig:comparisons_p_reset_supp}
\end{subfigure}%
\begin{subfigure}
  \centering
  \includegraphics[width=0.49\linewidth]{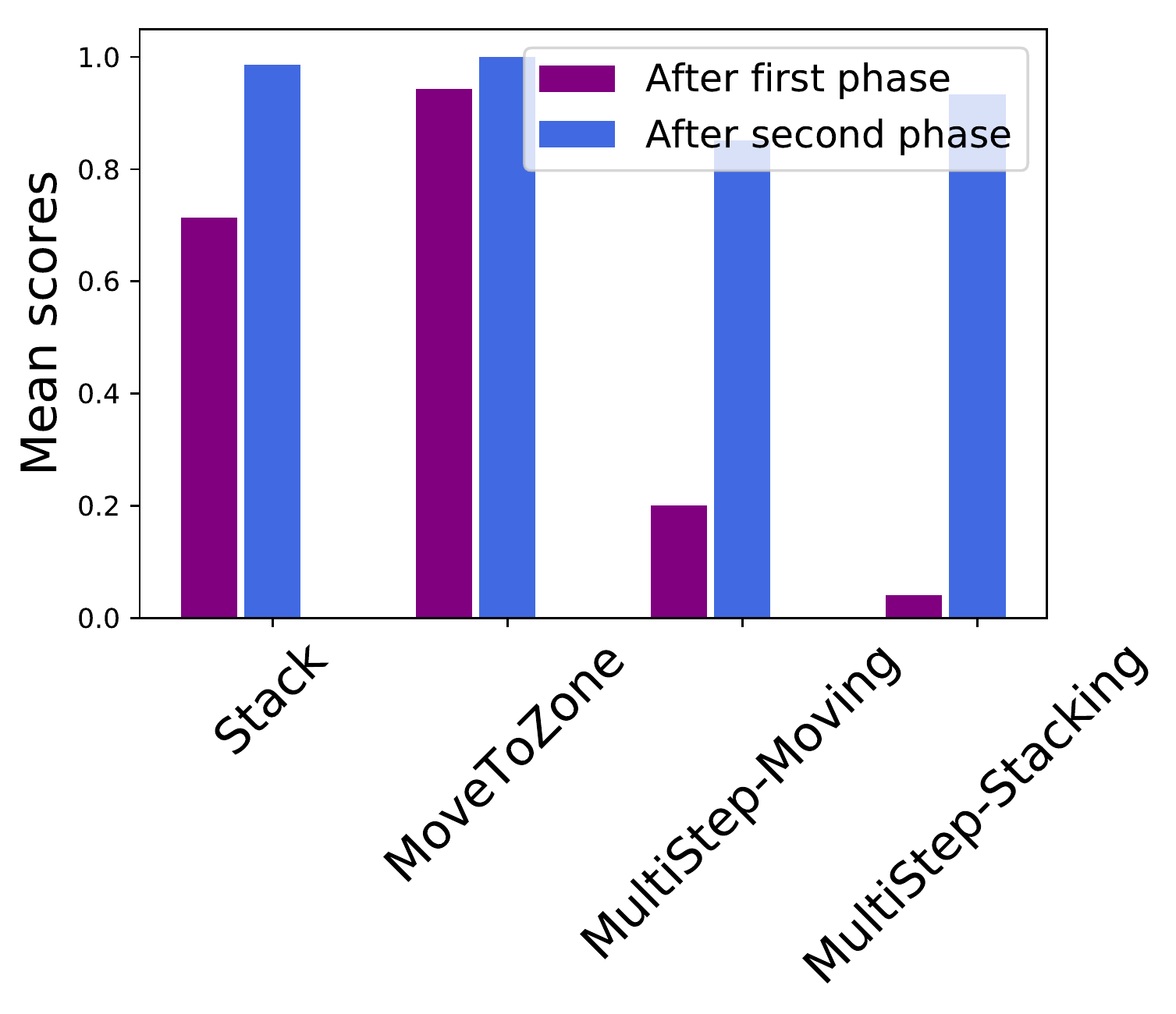}
  
\end{subfigure}
\caption{
\textbf{Left} We compare the goal-conditioned policy's performance on executing atomic programs {\sc stack} and {\sc movetozone} as well as two multi-step sequential atomic program executions after the completion of second phase with different probabilities $p$. \textbf{Right} We compare the performance of goal-conditioned policy in the first phase compared to the second phase. While the agent achieves good performance on atomic programs, the second phase is required to be able to sequence them (as observed in performance on multi-step tasks). Also, the reset probability can affect performance on multi-step tasks. We observe that $p=0.5$ gives the better trade-off between the asymptotic training performance and the agent ability to execute programs sequentially.\label{fig:comparisons_phases1_2}}
\end{figure}

We train the goal-conditioned policy $\atomp$ in two consecutive phases. In the first phase we sample the initial states and goal-vectors randomly. During the second phase, we change the initial and goal state distributions to ensure that our goal-conditioned policy still performs well when called sequentially by the meta-controller.  For initial states, with probability $p$ we do not reset the initial state of the environment between episodes. This means that the initial state of some episodes are the final state achieved in the previous episode. To simplify this process in a distributed setting, we store the final states of all episodes in a buffer and in each episode, with probability $p$, we sample a state from this buffer to be our initial state in this episode. Additionally, instead of random goal-vector sampling we randomly select an atomic program $p_i$, then given the initial state we compute the goal vector using the goal-setter. We study the impact of the reset probability $p$ as well as the usefulness of the second phase in \figureautorefname{}~\ref{fig:comparisons_phases1_2}.

\subsection{Results of self-behavioural model training}
\label{sec:supp_apem_learn}

We represent the self-behavioural model $\abm$ with a 2 layer MLP that takes as input the initial environment state and the atomic program and predicts the environment state after $T=50$ steps. Learning this model enables us to imagine the state of the environment following an atomic program execution and hence to avoid any further calls to atomic programs that would each have to perform many actions in the environment.

\begin{figure}[ht]
\vskip 0.2in
\begin{center}
\centerline{\includegraphics[width=0.5\textwidth]{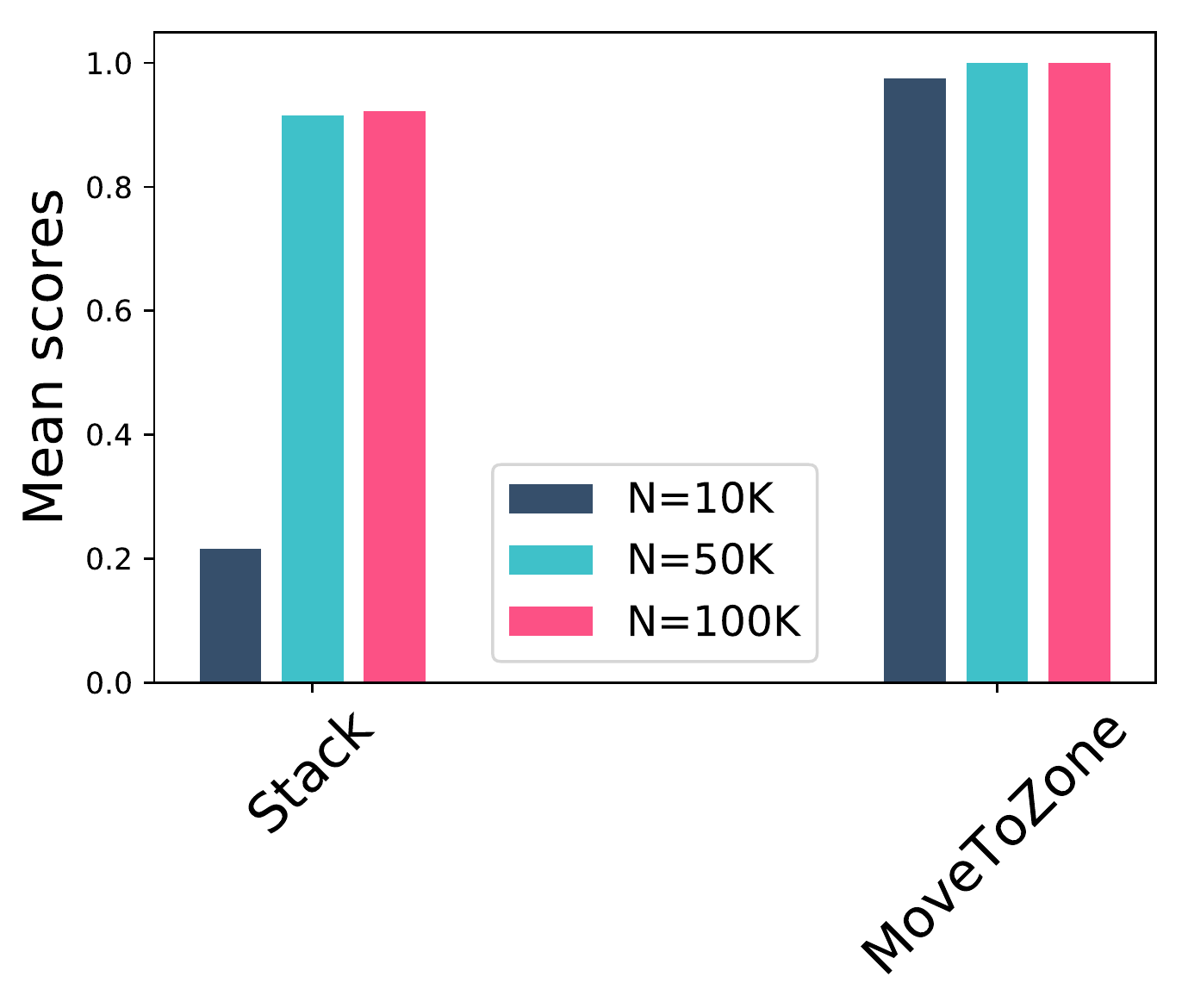}}
\caption{To train the self-behavioural model $\abm$, we construct a dataset. We record initial and final states for $N$ episodes of goal-conditioned policy's behaviour after training. At the beginning of each episode, an atomic program $p_i$ is randomly chosen. In this figure, we study the impact of the number of episodes $N$. }
\label{fig:apem_model_study_supp}
\end{center}
\vskip -0.2in
\end{figure}

To train the model, we play $N$ episodes with goal-conditioned policy in the environment on uniformly sampled atomic programs and record initial and final environment states in a dataset. Then, we train the model to minimize its prediction error, computed as a mean square error over this dataset. We study the impact of $N$ in \figureautorefname{}~\ref{fig:apem_model_study_supp}.

\subsection{Curriculum learning comparisons}

\begin{figure*}[ht!]
\vskip 0.2in
\begin{center}
\centerline{\includegraphics[width=0.9\textwidth]{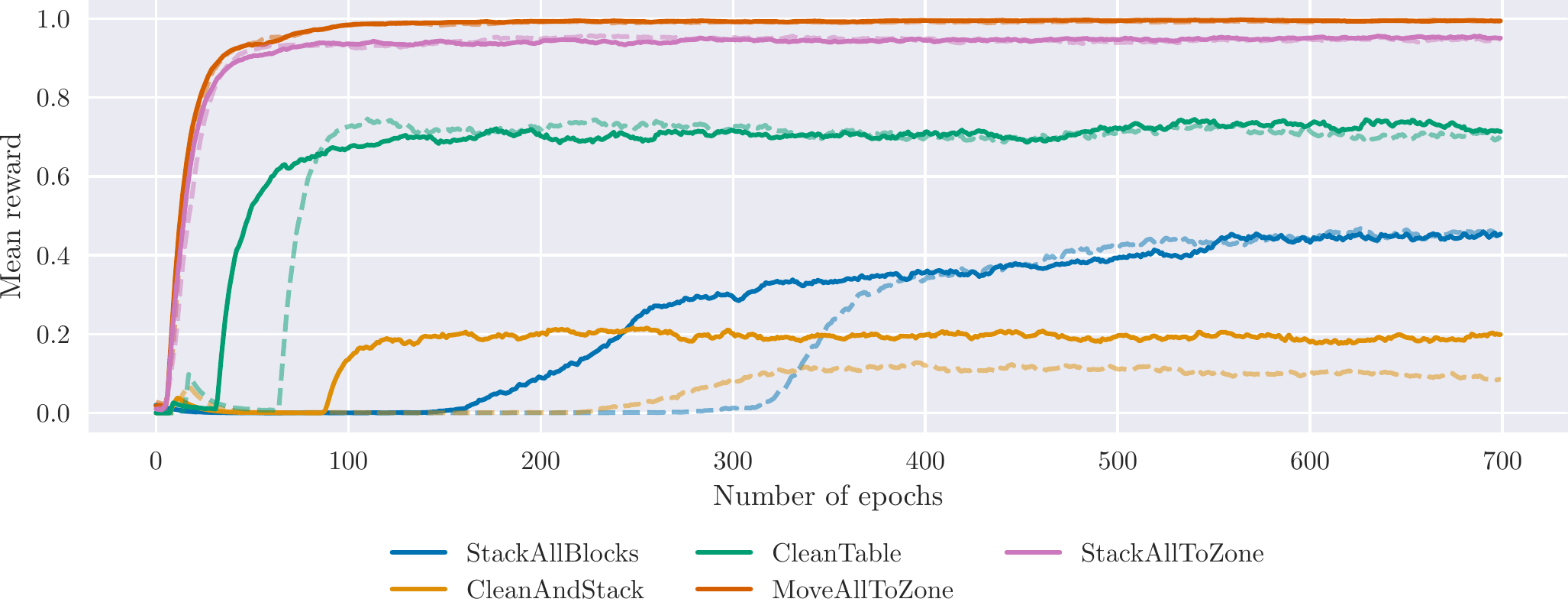}}
\caption{We represent the evolution of \anx performance during training and compare two curriculum strategies. Solid lines correspond to an agent trained with randomly sampled programs while dotted lines correspond to the programs  selected according to a learning progress based automatic curriculum.}
\label{fig:metacontroller_training_evolution_comparison}
\end{center}
\vskip -0.2in
\end{figure*}

We compare training the meta-controller with random program sampling compared to an automated curriculum using a learning progress signal detailed in \sectionautorefname{}~\ref{sec:supp_alphanpi_improvements}. We observe in \figureautorefname{}~\ref{fig:metacontroller_training_evolution_comparison} that both random and automated curriculum gives rise to a natural hierarchy for programs.  We observe that the {\sc MoveAllToZone} and {\sc StackAllToZone} are learned first. The traces generated by \anx after training confirms that these programs are used to execute the  more complex programs {\sc CleanAndStack} and {\sc CleanTable}. Given the results shown in \figureautorefname{}~\ref{fig:metacontroller_training_evolution_comparison} we cannot conclude that the learning progress based curriculum outperforms random sampling.

\subsection{Sample efficiency}
\label{sec:supp_comparisons}
We compare our method's sample efficiency with other works on the Fetch environment. For training the goal-conditioned policy, 28 \ddpg actors sample 100 episodes per epoch in parallel. Thus, 2800 episodes are sampled per epoch. In total, we train the agent for 250 epochs resulting in $7e^5$ episodes. To train the self-behavioural model, we show that sampling $5e^4$ episodes is enough to perform well. Thus, in total we only use $7.5e^5$ episodes to master all tasks. Note that training the meta-controller does not require any interaction with the environment. In comparison, in the \her original paper \citep{HER}, each task alone requires around $1.6e^5$ episodes to be mastered. In \curious \citep{curious}, training the agent requires $3e^5$ episodes. Finally, in \citep{curiosity_driven_multi_criteria} where the agent is trained only to perform the task {\sc Stack\_All\_Blocks}, around $2.5e^6$ episodes have been sampled. 

\end{document}